\documentclass[
]{ceurart}

\usepackage{listings}
\usepackage{booktabs}
\usepackage{geometry}
\usepackage{adjustbox}

\begin{document}

\copyrightyear{2025}
\copyrightclause{Copyright for this paper by its authors. Use permitted under Creative Commons License Attribution 4.0 International (CC BY 4.0).}
\conference{ISWC 2025 Companion Volume, November 2--6, 2025, Nara, Japan}

\title{DBLPLink 2.0 - An Entity Linker for the DBLP Scholarly Knowledge Graph}

\author[1]{Debayan Banerjee}[%
orcid=0000-0001-7626-8888,
email=debayan.banerjee@leuphana.de,
]
\cormark[1]
\address[1]{Leuphana University of Lüneburg, Lüneburg, Germany}
\address[2]{University of Hamburg, Hamburg, Germany}

\author[1,2]{Tilahun Abedissa Taffa}[%
orcid=0000-0002-2476-8335,
email=tilahun.taffa@leuphana.de,
]

\author[1]{Ricardo Usbeck}[%
orcid=0000-0002-0191-7211,
email=ricardo.usbeck@leuphana.de,
]

\cortext[1]{Corresponding author.}

\begin{abstract}
  In this work we present an entity linker for DBLP's 2025 version of RDF-based Knowledge Graph. Compared to the 2022 version, DBLP now considers publication venues as a new entity type called dblp:Stream. In the earlier version of DBLPLink, we trained KG-embeddings and re-rankers on a dataset to produce entity linkings. In contrast, in this work, we develop a zero-shot entity linker using LLMs using a novel method, where we re-rank candidate entities based on the log-probabilities of the "yes" token output at the penultimate layer of the LLM. The demo can be accessed at \url{https://dblplink-2.skynet.coypu.org/}.
\end{abstract}


\maketitle

\section{Introduction and Related Work}

Entity Linking (EL) is a task in natural language processing (NLP) that involves mapping named entities mentioned in text to their unique identifiers in a knowledge graph (KG). For example, in the question "Where did Albert Einstein study?", the label "Albert Einstein" needs to be linked to the unique entity identifier Q937\footnote{\url{https://www.wikidata.org/wiki/Q937}} in the Wikidata KG~\cite{vrandevcic2014wikidata}. Various entity linking systems have been developed~\cite{Sevgili2022} for general-purpose KGs like Wikidata, as well as for specialized domains such as biomedical~\cite{french2023overview} or financial~\cite{elhammadi-etal-2020-high} knowledge graphs.

Scholarly KGs are a specific type of knowledge graph focused on bibliographic data related to academic publications, authors, institutions etc. Examples of well-known scholarly KGs include OpenAlex~\cite{openalex}, ORKG~\cite{orkg}, and DBLP~\cite{dblp}. In this work, we concentrate on the DBLP KG, which is specifically designed for the computer science domain and is consequently smaller than more comprehensive scholarly KGs. DBLP—originally short for Data Bases and Logic Programming—was created in 1993 by Michael Ley at the University of Trier, Germany [5]. 

For entity linking over DBLP, a system named Deola~\cite{deola} was able to link author entities to DBLP documents in 2016. Notably, this predated the availability of DBLP in th RDF format.  In 2022, we released DBLPLink~\cite{debayandblplink} (which we henceforth refer to as DBLPLink 1.0), an entity linker built for the initially released version of the DBLP KG. The DBLP KG schemas before 2024 were built primarily around two major entity types: Creator and Publication. Subsequently in June 2024, DBLP introduced\footnote{\url{https://blog.dblp.org/2024/06/14/the-dblp-knowledge-graph-major-extension-and-an-update-to-the-rdf-schema/}} 
a new entity type dblp:Stream which encompasses multiple sub-classes under the broad category of publication venues, for example, conferences, journals, series and repositories. 

Our initial thought was to retrain DBLPLink 1.0 on the new KG and produce DBLPLink 2.0. However, moving DBLPLink 1.0 to a new KG requires computing new KG embeddings for all entities, retraining the entity label span detector, and re-training the re-ranker. In light of recent approaches using LLM-based prompting and zero-shot methods, we decided to build a new architecture from scratch for DBLPLink 2.0. DBLPLink 2.0 is able to link Person and Publication entity types as before, and additionally, can also link Stream entity types. DBLPLink 2.0 can be accessed at \url{https://dblplink-2.skynet.coypu.org/}. The code and data used to build this demo can be accessed at \url{https://github.com/semantic-systems/dblplink-2.0}.
\vspace{-2mm}
\section{User Interface}

\begin{figure}[h]
    \centering
    \includegraphics[width=0.9\textwidth]{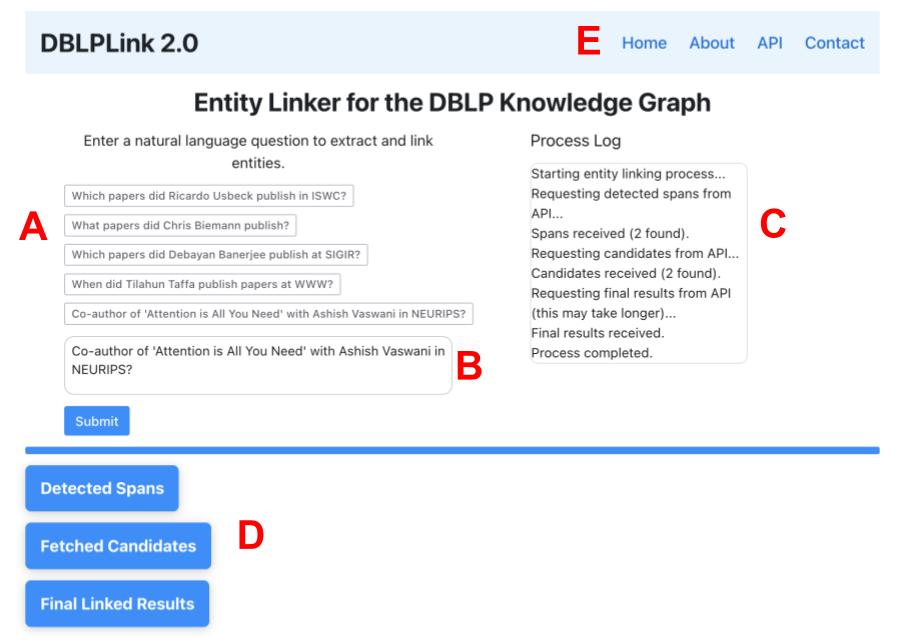}  
    \caption{The DBLPLink 2.0 Web Interface}
    \label{fig:master-ui}
\end{figure}
\vspace{-2mm}
As seen in Figure \ref{fig:master-ui}, the web UI is divided into five elements. \textbf{A} presents a set of question templates which maybe clicked and selected to fill up the text box. \textbf{B} carries the text box where the user may type an input text, and click on Submit to start the entity linking process. \textbf{C} displays a process log which is dynamically updated from the backend, keeping the user informed on the current step being executed. \textbf{D} is results area, where the detected mention spans and their types are displayed. Later, the fetched candidates form the text search are displayed. Finally the linked results are displayed under "Final Linked Results". \textbf{E} is a carousel of sub-pages, which provides further information, such as how to access the entity linker via an API call, more information about the backend entity linker architecture, and details of how to contact the authors and maintainers. 

\begin{figure}[h]
    \centering
    \includegraphics[width=0.9\textwidth]{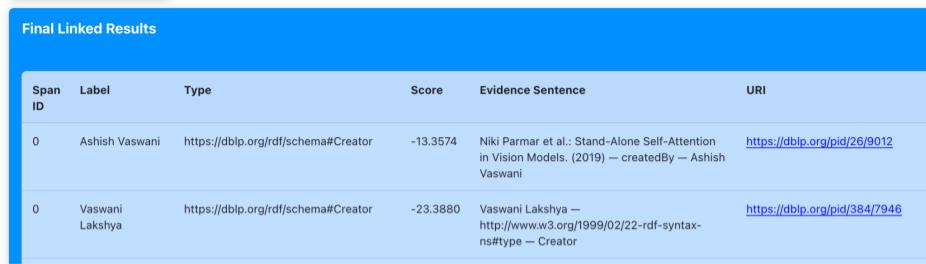}  
    \caption{The "Final Linked Results" Tab when expanded}
    \label{fig:results}
\end{figure}

Further, as seen in Figure \ref{fig:results}, the final linked results tab, when expanded, displays a sorted list of linked entities by log probability score, per span. The \textbf{first column} is the Span ID, where 0 stands for the first span, 1 stands for the second span and so forth. The \textbf{second column} is the entity label of the candidate as fetched from the Elasticsearch label database. The \textbf{third column} displays the DBLP type for the entity candidate. The \textbf{fourth column} displays the log probability score of the given entity. Note that the scores are in negative, and hence, they appear sorted in descending absolute value scores. The \textbf{fifth column} is also called the evidence sentence, which is the triple that produced the strongest log probability score for among all the triples for this given entity. The \textbf{sixth column} provides a clickable URL link for the entity, which takes the user to the entity's DBLP page.

\vspace{-2mm}
\section{Entity Linker Architecture}

Our entity linking pipeline combines prompted large language models (LLMs), type-specific retrieval from an Elasticsearch index, and neighborhood-based re-ranking using KG context. We illustrate the method using the input question:

\begin{quote}
\texttt{Who are the co-authors of Ashish Vaswani in "Attention is All You Need"  in neurips?}
\end{quote}
\vspace{-2mm}
\subsection{Mention and Type Extraction via Prompted LLM}
\vspace{-2mm}
We first extract named entity mentions from the input using a prompted LLM. The prompt is as follows:

\begin{quote}
\small
\texttt{
You are an information extraction assistant.\\
Extract named entities from the following sentence and classify them into one of the following types: person, publication, venue.\\
Let the output be a JSON array of objects with fields 'label' and 'type'.\\
Not all types may be present in a sentence. Now extract entities from the following sentence:\\
Sentence: "Who are the co-authors of Ashish Vaswani in the 'attention is all you need' paper in neurips?"\\
Entities:
}
\end{quote}

The LLM produces:

\begin{verbatim}
[
  {"label": "Ashish Vaswani", "type": "person"},
  {"label": "attention is all you need", "type": "publication"},
  {"label": "neurips", "type": "venue"}
]
\end{verbatim}
\vspace{-2mm}
\subsection{Candidate Entity Retrieval}

Each extracted label is matched against a type-specific Elasticsearch index to retrieve a list of candidate entities. For example:

\begin{itemize}
    \item \texttt{"Ashish Vaswani"} → [Ashish Vaswani, Vicky Vaswani, ...]
    \item \texttt{"attention is all you need"} → [doi:10.5555/attention-paper, ...
    \item \texttt{"neurips"} → [NeurIPS, NeurIPS 2022, NeurIPS 2023, ...]
\end{itemize}

\subsection{Knowledge Graph Neighborhood Expansion}

For each candidate entity, we fetch up to $N$ one-hop neighbors from a knowledge graph. These triples are converted into readable sequences using a template of the form:
\vspace{-2mm}
\begin{quote}
\texttt{[Head] - [Relation] - [Tail]}
\end{quote}
\vspace{-2mm}
Example for \texttt{Ashish Vaswani (author)}:
\vspace{-2mm}
\begin{itemize}
    \item \texttt{Ashish Vaswani - authored - attention is all you need}
    \item \texttt{Ashish Vaswani - affiliated with - Google Brain}
    \item \texttt{Ashish Vaswani - published at - NeurIPS}
\end{itemize}

This yields a set of short sentences describing the local graph structure of each candidate.

\subsection{Candidate Scoring with LLM Log-Probability}

Each linearized triple is evaluated by an LLM in the context of the original question. The prompt is:
\vspace{-2mm}
\begin{quote}
\small
\texttt{
Given this input text: "Who are the co-authors of Ashish Vaswani in the 'attention is all you need' paper in neurips?"\\
And the neighborhood context:\\
Ashish Vaswani - authored - attention is all you need\\
Is this the correct entity?\\
Answer with 'yes' or 'no'.}
\end{quote}
\vspace{-2mm}
We extract the log-probability of the next token being \texttt{"yes"} (before generation), which serves as a soft alignment score for that triple. Each candidate entity receives multiple such scores — one per triple. These are aggregated using {mean pooling, where the average log-probability over all triples is computed.

\subsection{Entity Re-ranking}

All candidate entities for a given mention are ranked according to their aggregated log-probability scores. The top-ranked candidate is selected as the final linked entity.
\vspace{-2mm}
\section{Implementation Details}
\vspace{-2mm}
The web demo is implemented using the Reflex web development framework\footnote{\url{https://reflex.dev/}} which allows building dynamic web interfaces written purely in Python. For finding optimal parameters for the different components of the entity linker pipeline, we randomly selected a set of 100 questions from the test set of the DBLP\_QuAD dataset~\cite{dblp_quad}. As seen in Table \ref{tab:experiments}, we tested several different LLMs of small sizes, keeping in mind the limited GPU infrastructure available to us as university based researchers. We tested \textit{0.5B, 1.5B, 3B, 7B, 14B} models of the \textit{Qwen-2.5} family and \textit{Llama-3.1-8B} and \textit{Mistral-7B-Instruct-v0.2}. Based on the results of our experiments, we found the Mistral model lagging far behind, with F1 score of 0.09. In comparison, Qwen-2.5-3B provided an optimal balance between size and performance, hence the web demo makes use of this model. The "text only" performance in the fourth row is a setting where the top text-based match is chosen as the final entity linking result. In effect, the subsequent neighbourhood-based re-ranking step is skipped. When comparing this result to the row above, it is clear that the entity linker is performing better than pure text-match-based entity linking. Additionally, from the last column's results, it seems that only for 62\% of the cases do the labels produced by the mention span detector translate to relevant candidates being fetched from the Elasticsearch labels database. All the experiments were performed with a setting of n=10 and k=10, where n=number of candidates from text search and k=number of neighbours from entities. We performed experiments with greater n and k, but saw negligible improvements when compared to the rise in execution time given the larger context to be parsed by the LLMs. Hence, we settled for values of 10 for n and k.
\vspace{-2mm}
\section{Limitations and Future Work}
\vspace{-2mm}
Due to non-availability of a new entity linking dataset over the current DBLP schema, we were unable to perform extensive evaluation for this task, especially on the new dblp:Stream entity type. Also, because the underlying KGs are different, we could not directly compare DBLPLink 2.0's performance with DBLPLink 1.0. As future, work, we shall prioritise the collection of a new dataset which would allow deeper analysis of our entity linker.

\begin{table}[htbp]
\centering
\caption{Evaluation Results on a test set of 100 questions from DBLP\_QuAD}
\label{tab:experiments}
\begin{adjustbox}{max width=\textwidth}
\begin{tabular}{@{}lccccc@{}}
\toprule
\textbf{Model / Setting} & \textbf{F1} & \textbf{MRR} & \textbf{Hits@1} & \textbf{Hits@5} & \textbf{Hits@10} \\
\midrule

  Qwen-0.5b   & 0.0000 & 0.0000 & 0.0000 & 0.0000 & 0.0000 \\
  Qwen-1.5b   & 0.2433 & 0.3721 & 0.3100 & 0.4400 & 0.4600 \\
  Qwen-3b     & \textbf{0.4400} & 0.5388 & \textbf{0.4900} & 0.5900 & 0.6200 \\
  Qwen-3b text only & 0.3867 & 0.4844 & 0.4300 & 0.5600 & 0.6200 \\
  Qwen-7b     & 0.3833 & 0.4919 & 0.4100 & 0.5900 & 0.6200 \\
  Qwen-14b    & 0.4300 & \textbf{0.5525} & 0.5000 & \textbf{0.6200} & 0.6200 \\
  LLaMA-3.1-8b  & 0.4000 & 0.4922 & 0.4000 & 0.6200 & 0.6400 \\
  Mistral-7b   & 0.0900 & 0.1841 & 0.1000 & 0.2600 & 0.4800 \\
\bottomrule
\end{tabular}
\end{adjustbox}
\end{table}

\vspace{-2mm}
\section{Declaration on the Use of Generative AI}
\vspace{-2mm}
No use of generative AI was made in writing this paper. We relied on the spell-check feature of Sharelatex software which was provided to us by the University of Leuphana as a tool to write research papers. ChatGPT was used for generating the initial templates of the code that the demo runs on. The code was later improved by the authors themselves to make it fully functional.

\bibliography{sample-ceur}

\end{document}